\documentclass[10pt,twocolumn,letterpaper]{article}

\usepackage[pagenumbers]{cvpr} 

\usepackage[dvipsnames]{xcolor}

\usepackage{multirow}
\usepackage{gensymb}
\usepackage{dblfloatfix}

\definecolor{cvprblue}{rgb}{0.21,0.49,0.74}
\usepackage[pagebackref,breaklinks,colorlinks,citecolor=cvprblue]{hyperref}

\def\paperID{393} 
\def\confName{3DV\xspace}
\def\confYear{2026\xspace}

\title{\textsc{Sat-Skylines}: 3D Building Generation from Satellite Imagery and \\ Coarse Geometric Priors}

\author{
Zhangyu Jin$^{1}$ \hspace{12pt} Andrew Feng$^{1}$\\
\\
$^1$ University of Southern California, Institute for Creative Technologies \\
{\tt\small \{zjin,feng\}@ict.usc.edu}
}

\begin{document}
\maketitle


\begin{figure*}[h]
\centering
\includegraphics[width=1\linewidth]{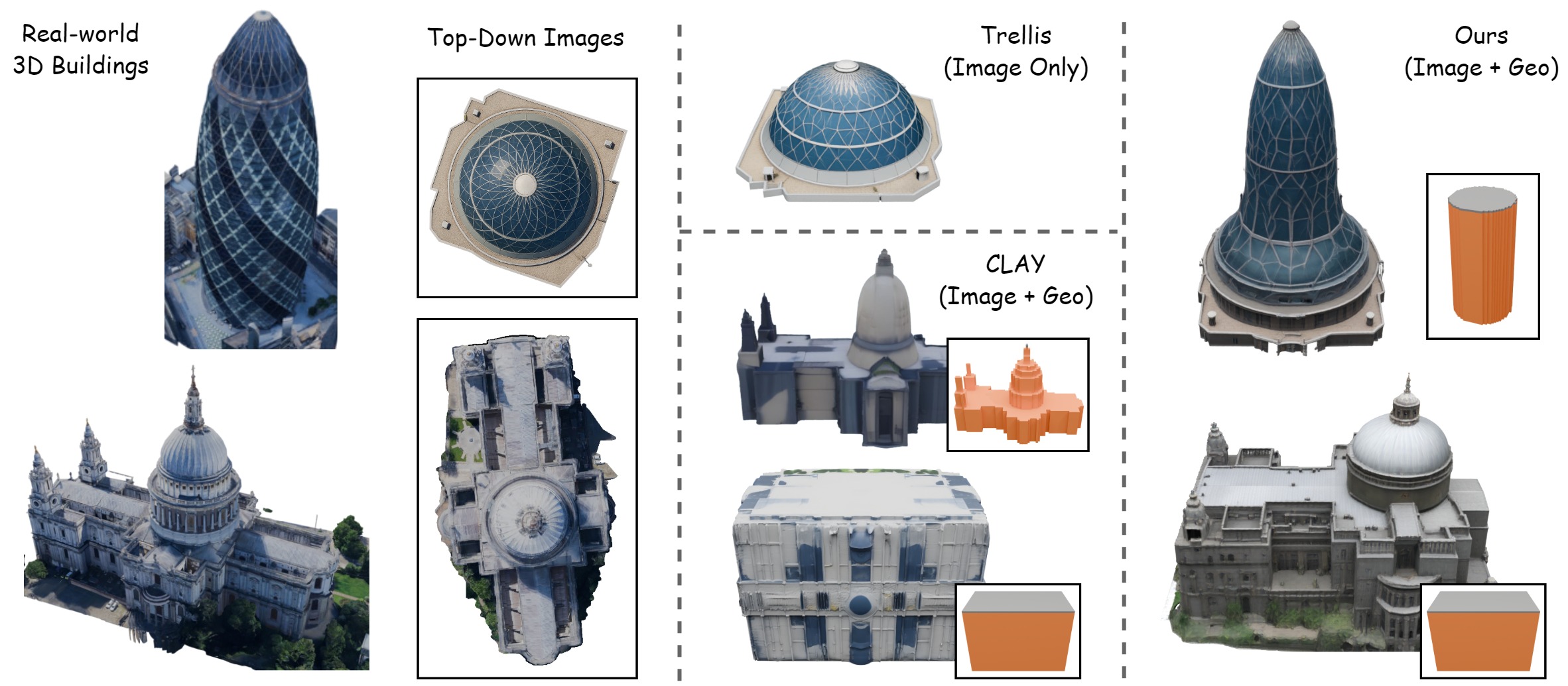}
\caption{\textbf{Necessity of our method}. Trellis fails to recover building heights (upper-middle). CLAY requires highly detailed voxels to work well (lower-middle). Our method takes top-down images and coarse geometric priors to generate realistic 3D buildings (right). }
\label{fig:ncessity_of_topic}
\end{figure*}

\begin{abstract}
We present \textbf{SatSkylines}, a 3D building generation approach that takes satellite imagery and coarse geometric priors.
Without proper geometric guidance, existing image-based 3D generation methods struggle to recover accurate building structures from the top-down views of satellite images alone. On the other hand, 3D detailization methods tend to rely heavily on highly detailed voxel inputs and fail to produce satisfying results from simple priors such as cuboids.
To address these issues, our key idea is to model the transformation from interpolated noisy coarse priors to detailed geometries, enabling flexible geometric control without additional computational cost. We have further developed \textbf{Skylines-50K}, a large-scale dataset of over 50,000 unique and stylized 3D building assets in order to support the generations of detailed building models. Extensive evaluations indicate the effectiveness of our model and strong generalization ability.

\end{abstract}

\section{Introduction}

Condtional 3D Building Generation is a growing research direction in 3D World Generation \cite{duan2025worldscore}, attracting attention from both academic and industry. Instead of relying on oblique aerial-view or street-view images, which are commonly used as the conditions for generating 3D models, we choose to leverage satellite imagery for its globally available large-scale data. This enables scalable 3D building modeling in regions where ground-level or oblique data is unavailable despite its lower resolution and fixed top-down view.


Although numerous models \cite{xiang2025structured, li2024craftsman3d,li2025triposg,he2025sparseflex,zhao2025hunyuan3d, lai2025hunyuan3d, hunyuan3d2025hunyuan3d,ye2025hi3dgen,wu2025direct3d,li2025step1x,gao2025mars, chen2024decollage, deng2024detailgen3d, chen2025art, chen2023shaddr, dong2024coin3d, zhang2024clay} have been proposed, achieving high-quality 3D building generation from top-down satellite images and coarse geometric priors remains challenging due to the following difficulties:

\textbf{Inefficient Usage of Satellite Images and Coarse Geometries}.
Many existing image-based 3D generation methods \cite{xiang2025structured, li2024craftsman3d,li2025triposg,he2025sparseflex,zhao2025hunyuan3d, lai2025hunyuan3d, hunyuan3d2025hunyuan3d,ye2025hi3dgen,wu2025direct3d,li2025step1x} use the Sphere Hammersley Sequence \cite{wong1997sampling} to render image prompts from 3D assets during training, but they typically ignore pure top-down views.
Top-down or satellite-view images are more challenging than traditional front or side views, because they provide no explicit height data and omit vertical surface details. In challenging cases, the lack of vertical cues in satellite imagery can result in ambiguous or inaccurate predictions, including failure to generate building-like structures.
  Consequently, such methods, like Trellis \cite{xiang2025structured}, often fail to reconstruct accurate building geometries using satellite imagery (see Fig.\ref{fig:ncessity_of_topic}).
Other methods support both image and voxel controls, but require highly detailed geometric voxels to achieve satisfactory results \cite{dong2024coin3d, zhang2024clay}.
As shown in Fig.~\ref{fig:ncessity_of_topic}, CLAY \cite{zhang2024clay} requires voxels to be detailed enough to do further refinement.
However, if provided with only a simple cuboid, it fails to add necessary geometric complexity.
Moreover, approaches \cite{dong2024coin3d, zhang2024clay} that support both image prompts and geometric priors suffer from prohibitively slow inference, limiting their potential applicability in large-scale building model generations.
To address these limitations, we design our method to efficiently leverage both top-down satellite imagery and flexible coarse geometry priors while maintaining fast inference speed.

\begin{figure*}[h]
\centering
\includegraphics[width=0.8\linewidth]{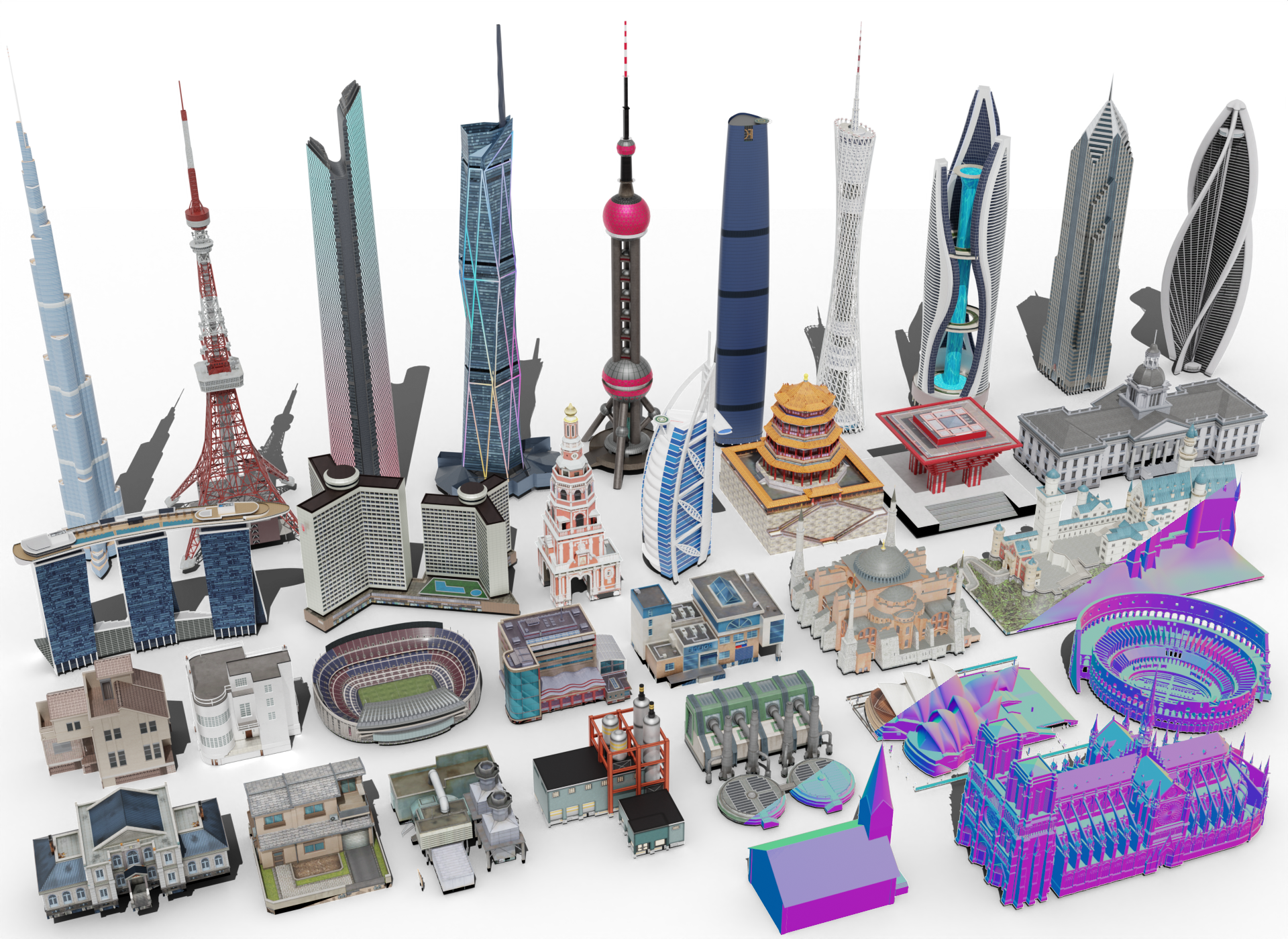}
\vspace{1.0em}
\caption{\textbf{Skylines-50K} is a large-scale, diverse, high quality 3D building dataset. These assets are sourced from the Steam Workshop of the famous city-building and simulation game `\textit{Cities: Skylines}'. Examples of rendered buildings are shown here to demonstrate style diversity.}
\label{fig:skylines_50k_overview}
\end{figure*}

\textbf{Lack of High-Quality 3D Building Datasets}.
Scale, diversity, and quality have been crucial factors in recent AI.
However, existing 3D building datasets \cite{rotterdam3d, digitaltwinvictoria, geoportalhamburg, geoportalleipzig, 3dmodelldresden, espoo3d, kuopio3d} often fall short in these aspects:
(\textit{a}) \textbf{Scale}. 
Although some datasets \cite{yang2023urbanbis, hu2024ss3dm, gao2025sum, xiong2023twintex} include collections of 3D building assets, they are in the form of colorless meshes, point clouds, or images. Datasets with textured meshes remain relatively small in scale.
(\textit{b}) \textbf{Diversity}. 
Certain datasets \cite{lin2022capturing, hua2025sat2city} focus on buildings from only a few cities (e.g., NYC, LA), lacking the architectural variety all over the world. 
This narrow coverage lowers their ability to in-the-wild settings.
In our work, we address these limitations by constructing a large-scale, globally diverse, and high-quality dataset of textured 3D building assets, enabling models to learn from both geometric detail and realistic appearance.


In this paper, we introduce \textbf{SatSkylines}, a 3D building generative approach that takes top-down satellite imagery together with coarse geometric priors as control.
Following Trellis \cite{xiang2025structured}, the top-down satellite image is injected into the sparse structure flow transformer as the condition.
Rather than feeding in pure noise (as in Trellis \cite{xiang2025structured}) or using just the coarse geometry prior (like DetailGen3D \cite{deng2024detailgen3d}), we apply a cosine interpolation between those two.
This process is further refined with a latent normalization on the coarse geometry priors to ensure both items follow the same gaussian distribution before interpolation.
This also allows controlling the intensity of the geometry priors as the constraints by simply changing the interpolation parameters to balance between higher fidelity and higher creativity.
By exploiting these techniques, our method can support both top-down satellite images and coarse geometry priors for generating high quality 3D building models.

We also propose \textbf{Skylines-50K}, a large-scale 3D building dataset containing over 50,000 unique and stylized assets (Fig.~\ref{fig:skylines_50k_overview}).
The dataset is sourced from the Steam Workshop of the popular city-building and simulation game `\textit{Cities: Skylines}'. 
Over the past 10 years, dedicated players all over the world have created and shared 3D building assets from their own countries, resulting in a rich collection that spans diverse architectural styles. The dataset includes landmarks, buildings of varying heights and scales, and distinctive structures, forming a valuable resource for developing high-quality, generalizable 3D building generation models.

\begin{figure*}[h]
\centering
\includegraphics[width=.9\linewidth]{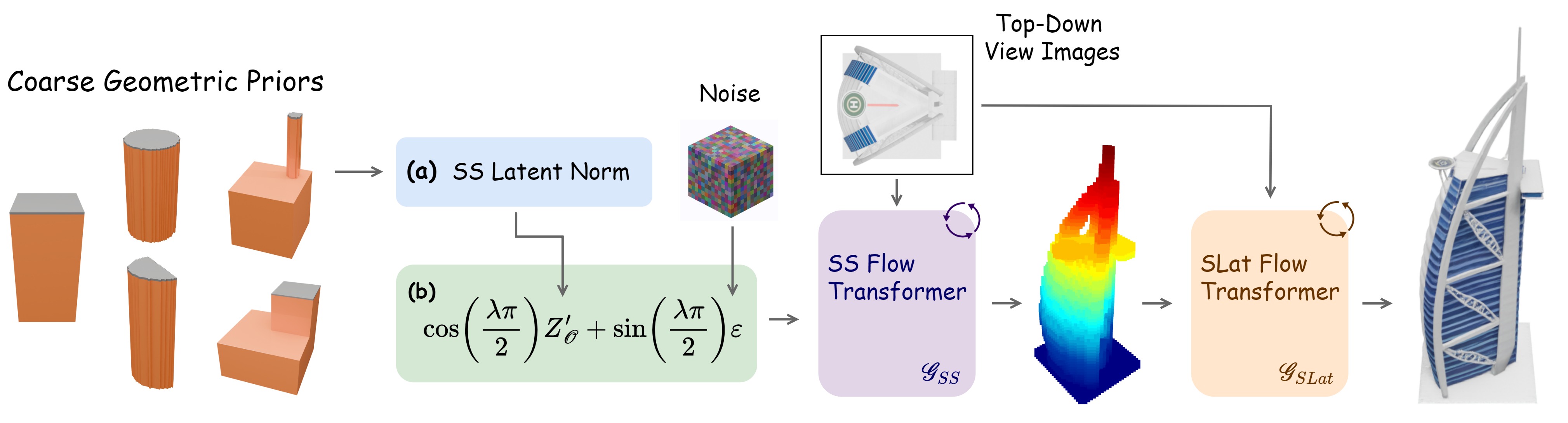}
\caption{\textbf{SatSkylines Architecture}. The coarse geometric prior $\mathcal{O}$ is encoded by the SS VAE to obtain $Z_{\mathcal{O}}$. (a) A channel-wise latent normalization is applied to produce $Z_{\mathcal{O}}^{'}$. (b) The cosine geometric interpolation is then performed between $Z_{\mathcal{O}}^{'}$ and gaussian noise $\epsilon$, with $\lambda$ controlling geometric guidance strength. Finally, the SS and SLat flow transformers generate detailed geometry and appearance.}
\label{fig:model_arch}
\end{figure*}

Based on the proposed model, we have also developed an end-to-end pipeline that generate 3D building assets from real world satellite imagery.
Given a GPS coordinate, our pipeline automatically gathers and enhances satellite imagery while generating coarse 3D geometry as priors. The prepared data is then fed into the generative model, providing a streamline process for real-world 3D building generation. We also evaluated our method on the Skylines-50K dataset and a diverse set of randomly selected real-world buildings. Our experiment results showed that SatSkylines is able to generate reliable outputs that align well with both the satellite imagery and the coarse geometry constraints. Our main contributions are summarized as follows.

(\textit{a}) A large-scale and high quality 3D building dataset.

(\textit{b}) An effective 3D building generation architecture using both satellite imagery and coarse geometry priors.

(\textit{c}) An end-to-end pipeline that produces detailed 3D building assets from real world satellite imagery.

\section{Related Works}

\textbf{3D Building Datasets}.
Existing 3D building datasets fall short in the following aspects:
(\textit{a}) Non-textured 3D format. 
Some provides only colorless meshes, including Building3D \cite{wang2023building3d} and so on \cite{huang2022city3d, lee2025nuiscene, wenrealcity3d}. 
Others are available only as point clouds, including SensatUrban \cite{hu2022sensaturban} and others \cite{chen2022stpls3d, wei2023buildiff, yin2025archidiff}. 
A large number of them are only composed of images, such as CityTopia \cite{xie2025citydreamer4d} and some else \cite{xie2024citydreamer, li2023matrixcity, turki2022mega, wang2024xscale, kerbl2024hierarchical, yao2020blendedmvs, liu2023deep, xiong2024gauu, jiang2025horizon, zhang2024drone}.
Additionally, BuildingNet \cite{selvaraju2021buildingnet} has colorized 3D buildings but with non-photorealistic colors.
(\textit{b}) Small at scale. 
Existing datasets with good textured 3D buildings have limited number of assets, such as UrbanBIS \cite{yang2023urbanbis}, SS3DM \cite{hu2024ss3dm}, SUMParts \cite{gao2025sum}, and TwinTex \cite{xiong2023twintex}. 
(\textit{c}) Lack of diversity. 
Current large-scale textured 3D building datasets usually focus on only a few cities (e.g., NYC, LA), lacking the architectural variety all over the world, such as UrbanScene3D \cite{lin2022capturing} and Sat2City \cite{hua2025sat2city}.
(\textit{d}) Quality constraints.
Many recent large-scale, diverse datasets are not fully handcrafted, resulting in bumpy surfaces and low-resolution textures, such as Google Earth \cite{lisle2006google} and other digital twin cities datasets \cite{rotterdam3d, digitaltwinvictoria, geoportalhamburg, geoportalleipzig, 3dmodelldresden, espoo3d, kuopio3d}.
In contrast, our Skylines-50K dataset offers over 50,000 handcrafted and textured 3D building assets from diverse locations worldwide.

\textbf{3D Generation Control Types}.
High-resolution image-based 3D generation models have advanced rapidly, such as Trellis \cite{xiang2025structured} and others \cite{li2024craftsman3d,li2025triposg,he2025sparseflex,zhao2025hunyuan3d, lai2025hunyuan3d, hunyuan3d2025hunyuan3d,ye2025hi3dgen,wu2025direct3d,li2025step1x}.
But they are neither trained on top-down satellite imagery nor designed to support geometric control.
Conversely, some geometry-controlled 3D generation models lack the ability to take image prompts. 
For instance, Cube \cite{bhat2025cube} supports text prompts with 3D bounding box control, and Sat2City \cite{hua2025sat2city} uses only height maps.
3D detailization methods refine coarse geometric priors into detailed geometry but also face limitations.
Mars \cite{gao2025mars} and DECOLLAGE \cite{chen2024decollage} only accept coarse geometry without image inputs, while DetailGen3D \cite{deng2024detailgen3d} supports both image prompts and coarse voxels but outputs geometry without texture or color.
ART-DECO \cite{chen2025art} can produce textured 3D assets but requires additional text prompts.
ShaDDR \cite{chen2023shaddr} and Coin3D \cite{dong2024coin3d} accept both image and coarse geometry inputs and output textured assets, but they demand unrealistically detailed input to achieve satisfactory results.
For example, when provided only with a simple cuboid, their outputs are of low quality. 
Rodin, a commercial product built on CLAY \cite{zhang2024clay}, supports diverse inputs including images, text, bounding boxes, voxels, and point clouds, and can generate high-quality textured assets.
However, it requires around 5–10 minutes per forward pass, making it impractical for large-scale automated use.
In comparison, our SatSkylines model efficiently processes top-down satellite imagery together with flexible coarse geometric priors to produce textured, detailed 3D building assets in around 15 seconds.

\textbf{3D Generation Control Mechanisms}.
Images are usually encoded using DINOv2 \cite{oquab2023dinov2} or CLIP \cite{radford2021learning} to extract visual features, which are then injected as key–value pairs into the cross-attention layers of a DiT \cite{peebles2023scalable}, as illustrated in Trellis \cite{xiang2025structured} and related works \cite{wu2025direct3d, li2025triposg}. 
In some cases, image features are concatenated directly with the noise latent and fed into the DiT, as in \cite{zhao2025hunyuan3d, hunyuan3d2025hunyuan3d, li2025step1x}.
We adopt the Trellis design \cite{xiang2025structured} for its strong trade-off between generation quality and computational efficiency.
For 3D geometry control, LION \cite{vahdat2022lion} proposes a training-free diffuse–denoise approach for point clouds control, but the absence of training also limits its generalization ability.
CLAY \cite{zhang2024clay} and others \cite{hua2025sat2city, dong2024coin3d} incorporate additional ControlNet \cite{zhang2023adding} or Adapters \cite{mou2024t2i} for voxel control, which increases model complexity and slows inference.
Cube \cite{bhat2025cube} encodes 3D bounding box coordinates as text via the CLIP text encoder \cite{radford2021learning}, but this design is difficult to extend to more complex voxel control.
DetailGen3D \cite{deng2024detailgen3d} feeds coarse geometry directly into the DiT, directly modeling the transformation between coarse and fine geometry.
However, this is unsuitable for our setting, where the model must handle cases ranging from a simple cuboid to diverse building shapes guided by various image prompts. 
Such a “one-to-many” mapping increases the difficulty of Rectified Flow \cite{liu2022flow} learning.
Instead, our SatSkylines model, built on Trellis \cite{xiang2025structured}, interpolates between pure noise and coarse geometric priors as the DiT input, preserving high inference speed and enabling simultaneous image and geometry control.

\section{Method}



An overview of our \textbf{SatSkylines} model is shown in Fig.~\ref{fig:model_arch}.
Inspired by Trellis \cite{xiang2025structured}, we first uses a Sparse Structure (SS) Rectified Flow \cite{liu2022flow} transformer to produce geometry latents $Z_{ss}$, followed by a Structured Latent (SLat) transformer to generate appearance latents $Z_{slat}$.
Top-down view images are introduced through cross-attention layers.
However, Trellis relies only on noise inputs and does not incorporate necessary geometric controls.


\subsection{Coarse Geometric Control}

Existing research, such as \cite{deng2024detailgen3d}, demonstrated potential in directly modeling transformation between coarse geometric priors $Z_{\mathcal{O}}$ and detailed ones $Z_{ss}$. 
However, these methods \cite{gao2025mars, chen2024decollage, deng2024detailgen3d, chen2025art, chen2023shaddr, dong2024coin3d, zhang2024clay} require highly detailed $Z_{\mathcal{O}}$ to achieve satisfactory results $Z_{ss}$, making them unsuitable for our use case.
Our goal is to develop a model capable of generating detailed geometry from simple starting priors, such as a basic cuboid.
This challenge is a `one-to-many' mapping problem, where a single, simple coarse prior could lead to many different detailed outcomes, which is more difficult for the Rectified Flow to learn.
To address this, we take an interpolation between $Z_{\mathcal{O}}$ and gaussian noise $\epsilon$ as the input for our SS Flow Transformer.
This approach not only provides the model with geometric prior information but also effectively addresses the "one-to-many" problem by varying the starting point with different level of noises. Unlike other techniques \cite{zhao2025hunyuan3d, lai2025hunyuan3d, hunyuan3d2025hunyuan3d} that increase computational costs by concatenating conditions with noise latent, this method does not affect training or inference speeds.

As shown in Fig. \ref{fig:model_arch}, any coarse geometric priors are first converted into a $N^3$ grid voxels
$\mathcal{O}\in\{0,1\}^{N\times N\times N}$,
where $1$ indicates an occupied voxel and 0 means empty.
We then apply the SS VAE encoder to get the SS latent representation 
$Z_{\mathcal{O}}\in\mathbb{R}^{D\times D\times D\times C^{'}}$,
where $D$ is down-sampled spatial resolution and $C^{'}$ is the corresponding feature dimension.

\textbf{SS Latent Normalization}. 
After obtaining the $Z_{\mathcal{O}}$, we perform channel-wise normalization. 
\begin{equation*}
Z_{\mathcal{O}}^{'}=\frac{Z_{\mathcal{O}}-Mean(Z_{\mathcal{O}})}{Std(Z_{\mathcal{O}})}
\end{equation*}
In our experiments, the distribution of $Z_{\mathcal{O}}$ does not follow a standard gaussian distribution but roughly $\mathcal{N}(0,0.2)$.
Without normalization, it makes the model less responsive to the geometric priors, leading to poor geometric control. 

\textbf{Cosine Geometric Interpolation}.
We define the interpolated geometric latent as follows.
\begin{equation*}
Z_{\mathcal{O}}^{''}=\cos(\frac{\lambda \pi}{2})Z_{\mathcal{O}}^{'}+\sin(\frac{\lambda \pi}{2})\epsilon
\end{equation*}
where $\epsilon$ is the pure gaussian noise. 
Since $Z_{\mathcal{O}}^{'}$ and $\epsilon$ follow gaussian distributions, their linear combination remains gaussian, owing to the identity $\cos^2(\frac{\lambda \pi}{2})+\sin^2(\frac{\lambda \pi}{2})=1$.
Here $\lambda\in[0,1]$ controls the degree of coarse geometric guidance:
when $\lambda=0$, the interpolation fully reflects the geometry priors $(Z_{\mathcal{O}}^{''}=Z_{\mathcal{O}}^{'})$;
when $\lambda=1$, it ignores the geometry priors and use only noise.
At inference, $\lambda$ can be manually set to adjust the amount of geometric influence.

\subsection{Dataset Curation
\label{sec:dataset_curation}}
Existing 3D building datasets \cite{hu2022sensaturban, chen2022stpls3d, wei2023buildiff, yin2025archidiff, xie2025citydreamer4d, xie2024citydreamer, li2023matrixcity, turki2022mega, wang2024xscale, kerbl2024hierarchical, yao2020blendedmvs, liu2023deep, xiong2024gauu, jiang2025horizon, zhang2024drone, yang2023urbanbis, hu2024ss3dm, gao2025sum, xiong2023twintex, lin2022capturing, hua2025sat2city, lisle2006google, rotterdam3d, digitaltwinvictoria, geoportalhamburg, geoportalleipzig, 3dmodelldresden, espoo3d, kuopio3d} are often limited in scale, diversity, or quality.
To address these gaps, we introduce a new dataset, \textbf{Skylines-50K}, as illustrated in Fig.~\ref{fig:skylines_50k_overview}.
We collect raw asset files from the Steam Workshop of `\textit{Cities: Skylines}' and convert them into meshes with PBR materials. All of the assets are then exported into the GLB format.

For each 3D asset, we render eight image conditions around the top view, ensuring that one of them is a pure top view.
Since the Skylines-50K dataset does not provide OpenStreetMap data, we generate coarse geometric priors ourselves.
Each asset is assigned three levels of priors (LOD 0–2): LOD 0 corresponds to a bounding box or cuboid, LOD 1 uses one unique cross-section along the height, and LOD 2 employs two distinct cross-sections. This design is intended to mimic OpenStreetMap building data, which typically contains 2D footprints and height attributes. During training, these four geometric priors together with a pure Gaussian noise are uniformly sampled within each batch.

\begin{figure*}[h]
\centering
\includegraphics[width=1\linewidth]{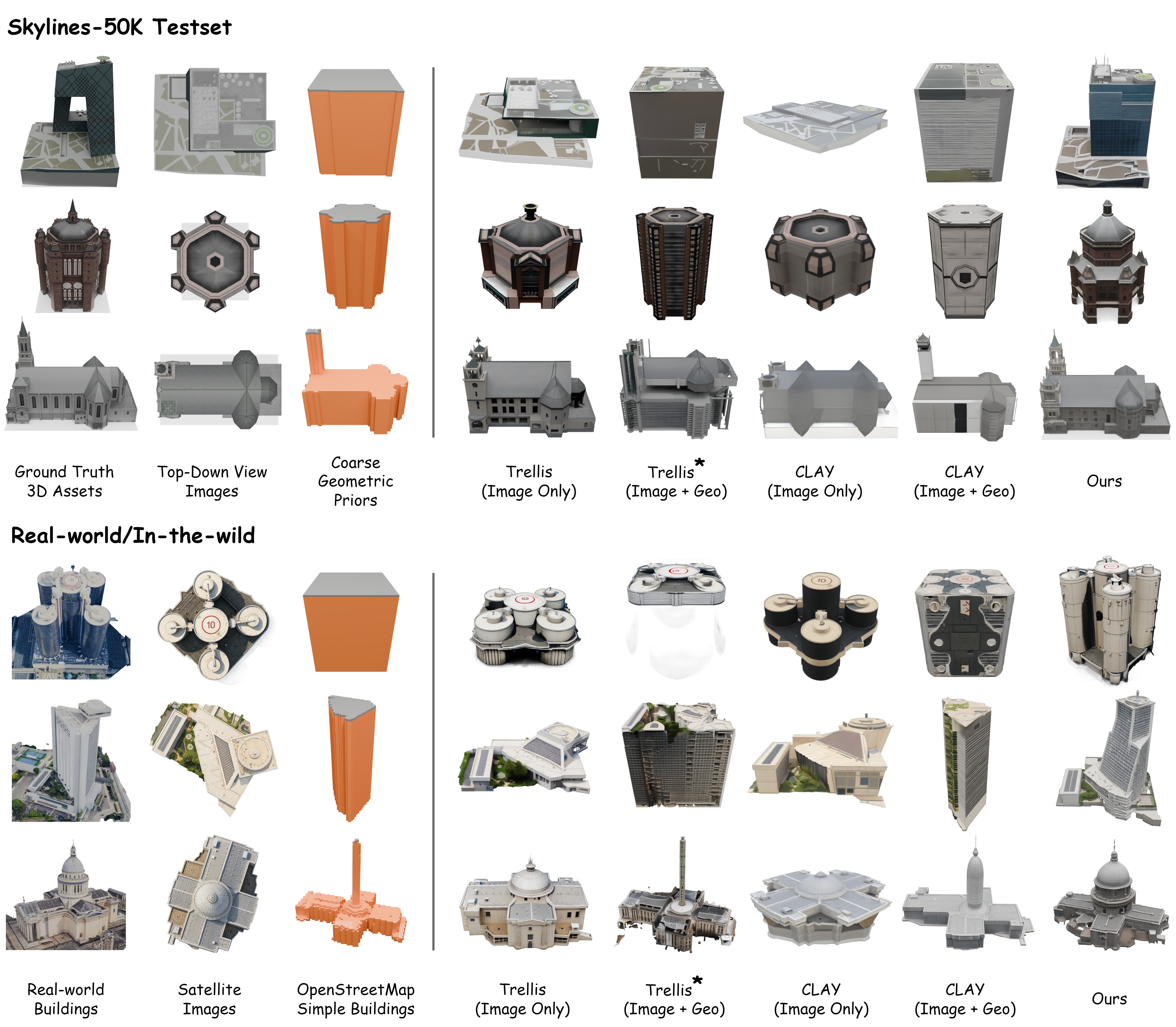}
\caption{Visual comparisons of generated 3D building assets between our method and previous approaches. The first three rows are from the Skylines-50K test set, and the last three rows are real-world examples. Here $\star$ indicates that Trellis does not natively support coarse geometric control, but our cosine geometric interpolation can be applied to incorporate geometric priors into its input. }
\label{fig:main_comparison}
\end{figure*}

\subsection{Real-world 3D Building Generation Pipeline \label{sec:realworld_pipeline}}

Image-based 3D generation methods \cite{xiang2025structured, li2024craftsman3d,li2025triposg,he2025sparseflex,zhao2025hunyuan3d, lai2025hunyuan3d, hunyuan3d2025hunyuan3d,ye2025hi3dgen,wu2025direct3d,li2025step1x} require users to provide an image prompt with foreground masks. 
In practice, Rembg \cite{gatis2023rembg} is usually applied to separate the foreground building from the background automatically. 
However, this approach tends to perform poorly on satellite images, since most background-removal models are not trained for this specific domain and fail to create clean masks. In addition, geometry-controlled 3D generation models \cite{gao2025mars, chen2024decollage, deng2024detailgen3d, chen2025art, chen2023shaddr, dong2024coin3d, zhang2024clay} require users to give a geometric prior. Typically, this involves creating voxel representations using software such as Mesh Editor \cite{Hyper3dMeshEditor}, Blender \cite{Blender}, or Unreal Engine \cite{UnrealEngine}. This presents a steep learning curve for both novice users and experienced artists to prepare the inputs for large scale building model generations. 


To address these issues and to demonstrate the proposed method, we have developed an end-to-end pipeline that requires only a geo-spatial bounding box of (min-lat, max-lat, min-lon, max-lon) to simplify the data preparation and building model generations. Specifically, from the bounding box information, it automatically retrieves 2D building footprints and height attributes from OpenStreetMap to generate coarse geometric priors, avoiding manual voxel creation. 
Satellite imagery from sources like Google Maps or Mapbox can also be combined with OpenStreetMap outlines to directly extract building masks without additional segmentation tools. Finally, since satellite imagery is often blurry or low-resolution, the pipeline also leverages `\textit{gpt-image-1}' to enhance the image quality with super-resolution.

\section{Experiments}

\begin{table}[]
\footnotesize
\setlength\tabcolsep{2pt}

\makebox[\columnwidth][c]{
\centering
\begin{tabular}{c|c|ccc|ccc}
\hline
\multirow{2}{*}{Methods} & \multirow{2}{*}{\# Assets} & \multicolumn{3}{c|}{Geometry}       & \multicolumn{3}{c}{Appearance} \\
                      &   & IoU $\uparrow$    & CD $\downarrow$ & F Score $\uparrow$ & PSNR $\uparrow$     & LPIPS $\downarrow$    & CLIP $\uparrow$   \\ \hline
Trellis \cite{xiang2025structured}              & 500   & 0.4415 & 0.0632           & 0.6295  &    11.9705      &     0.4529      &   0.6461      \\
CLAY \cite{zhang2024clay}                 & 20$^\star$  &   0.6859     &   0.0574               & 0.6516        &  13.3745        &   0.4136        &   0.6549      \\ \hline
\rule[-5pt]{0pt}{14pt} \multirow{2}{*}{\textbf{SatSkylines}}      &  500      & \textbf{0.9381} & \textbf{0.0168}           & \textbf{0.8684}  &   \textbf{19.2386}       &     \textbf{0.1678}      &     \textbf{0.7860}    \\
      &  20$^\star$      & 0.9278  &  0.0080        &  0.9091 &  18.6003        &  0.1786         &  0.7958       \\\hline
\end{tabular}
}
\caption{\footnotesize \textbf{Quantitative Comparisons in Skylines-50K Dataset}. Trellis uses image prompts only, while CLAY and our SatSkylines additionally use LOD 1 coarse geometric priors. Here $^\star$ means, due to CLAY’s slow inference speed, we evaluate it only on a smaller sub-test set of 20 assets.
}
\label{tab:skylines_50k}
\end{table}

\textbf{Implementation details}.
Our model is built upon Trellis with the 1.1B image-large configuration.
We initialize from Trellis pretrained weights and finetune on Skylines-50K using a learning rate of $1e^{-4}$, batch size of $32$, and $40K$ total steps on 4 A100 GPUs.
Channel-wise mean and standard deviation of $Z_{\mathcal{O}}$ are fixed constants for both training and inference.
The interpolation factor $\lambda$ follows a logit-normal distribution with parameters $\mu=1,\sigma=1$ during training, while being fixed to $0.5$ at inference.
At inference, classifier-free guidance (CFG) scales are set to 7.5 for SS and 3.0 for SLat, with 50 sampling steps for both modules.

\subsection{3D Building Generation}

In this section, we evaluate our 3D building generation quality. We first present various 3D generation results of our method, and then compare with other baseline methods.

\textbf{Skylines-50K Dataset}. 
Our evaluation is conducted on the Skylines-50K test set of 500 representative instances.
For the geometry evaluation, we empoly Chamfer Distance (CD), IoU, and F-score of generated sparse structure voxels.
For appearance quality, each instance is rendered into four views with FoV $40^{\degree}$, radius $2$, pitch $30^{\degree}$, and yaw angles $\{0^{\degree}, 90^{\degree}, 180^{\degree}, 270^{\degree}\}$.
We then compute PSNR, LPIPS, and CLIP similarity against ground truth images.
Following \cite{chen2025art}, in order to enable fair comparisons with Rodin (the commercial version of CLAY), we additionally define a 20-instance sub-test set. We conduct the comparison with CLAY on this sub-test set, as Rodin is both costly and requires approximately 5–10 minutes to generate a single building, making large-scale automatic evaluation impractical.

\begin{figure}[h]
\centering
\includegraphics[width=\linewidth]{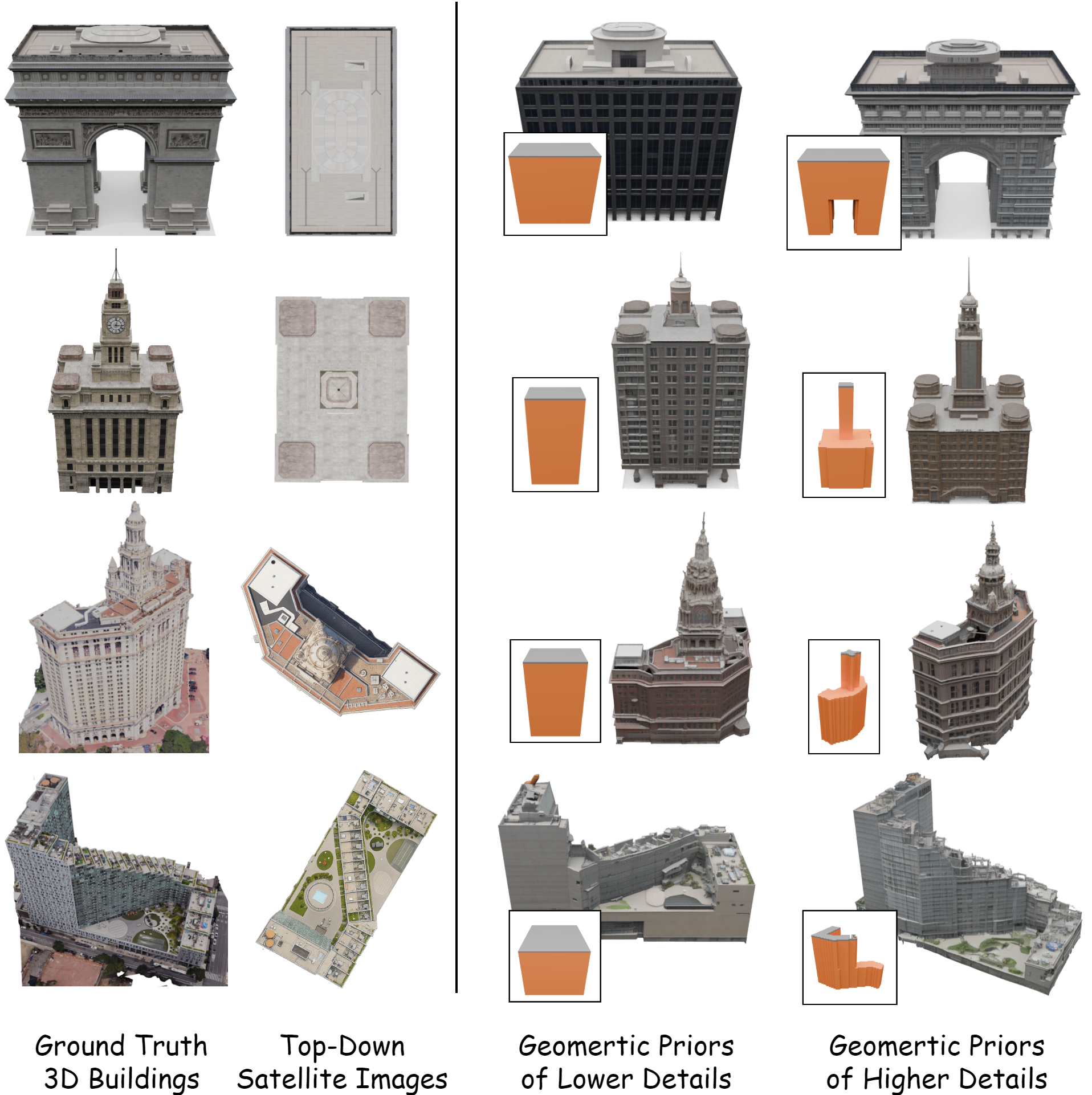}
\caption{\footnotesize \textbf{Visualization of Coarse Geometric Prior Variations}. 
The first two rows are examples from the Skylines-50K dataset, while the last two rows are real-world cases.
}
\label{fig:coarsess_control_comparison}
\end{figure}

\textbf{Real-world 3D Building Generation}.
We have also evaluated our model qualitatively in real-world, in-the-wild settings. 
Since no ground truth 3D assets are available, the qualitative results are provided with Google Earth screenshots as reference.

As shown in Fig.~\ref{fig:main_comparison}, our method produce better quality 3D building models compared to previous approaches, offering not only top-down satellite image control but also more precise alignment with flexible coarse geometric priors. Tab.~\ref{tab:skylines_50k} further demonstrates that our method shows competitive performance across all evaluation metrics.
Trellis \cite{xiang2025structured} generates reasonably good 3D buildings from satellite images but fails to capture realistic building geometries.
Trellis \cite{xiang2025structured} can also use our cosine geometric interpolation to accept geometric controls. However, the performance is not as reliable since it is not trained on such distributions with top-down view conditions, as shown in Fig.~\ref{fig:main_comparison} and Tab. \ref{tab:coarseness_control}.
CLAY \cite{zhang2024clay} relies heavily on voxel priors and lacks further refinement under satellite prompts.

\begin{table}[]
\small
\setlength\tabcolsep{4pt}

\makebox[\columnwidth][c]{
\centering
\begin{tabular}{c|c|ccc}
\hline
\multirow{2}{*}{Methods}              & \multirow{2}{*}{\begin{tabular}[c]{@{}c@{}}Geormtric Priors\\ Coarseness Level\end{tabular}} & \multicolumn{3}{c}{Geometry}                         \\
                                      &                                                                                              & IoU $\uparrow$            & CD $\downarrow$ & F Score  $\uparrow$       \\ \hline
Trellis  \cite{xiang2025structured} & LOD 2                                                                          &  0.6367         &  0.0529          &   0.6495        \\ \hline
\multirow{3}{*}{\textbf{SatSkylines}} & LOD 0                                                                          & \textbf{0.9515}          & 0.0222           & 0.8315          \\
                                      & LOD 1                                                                                         & 0.9381          & 0.0168           & 0.8684          \\
                                      & LOD 2                                                                                         & 0.9441 & \textbf{0.0141}  & \textbf{0.8943} \\ \hline
\end{tabular}
}
\caption{\footnotesize \textbf{Quantitative Analysis of Coarse Geometric Prior Variations}.
Trellis does not natively support coarse geometric control, but our cosine interpolation enables it to use geometric priors.
LOD 0 denotes a simple cuboid, while LOD 2 specifies two distinct cross-sections along the height.
}
\label{tab:coarseness_control}
\end{table}

\begin{figure}[h]
\centering
\includegraphics[width=\linewidth]{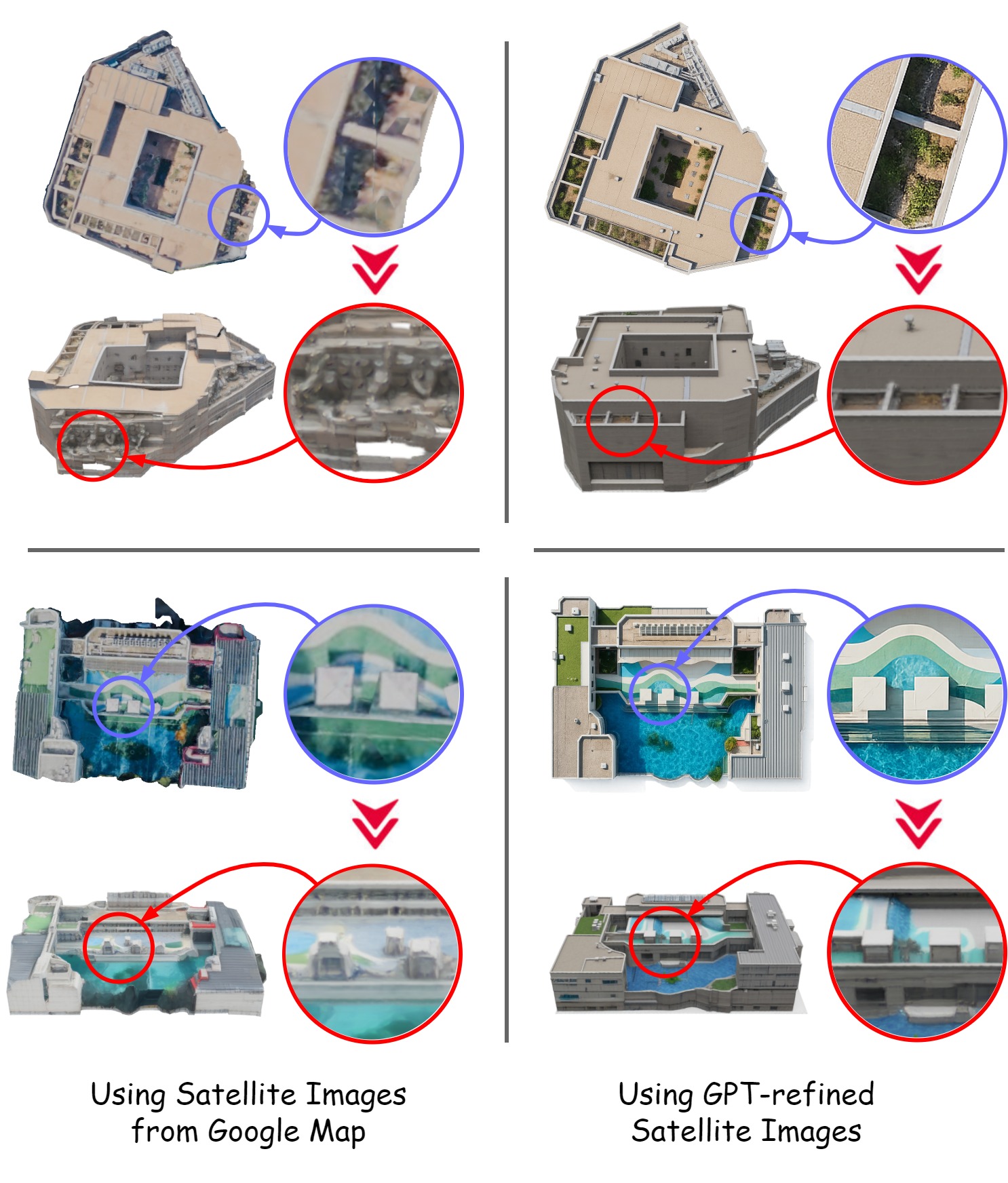}
\caption{\footnotesize \textbf{Visualization of GPT Satellite Image Refinement}.
Blue circles highlight zoomed-in details from 2D satellite images, while red circles mark the corresponding regions in SatSkylines generated 3D buildings. All examples are randomly sampled from real-world data.
}
\label{fig:gpt_refine_comparison}
\end{figure}

\subsection{Coarse Geometric Prior Variations}

SatSkylines can generate reasonable 3D buildings from very coarse priors, such as a simple cuboid. 
When provided with more detailed priors, the generated geometry further improves, producing more accurate and faithful building shapes.
As illustrated in Fig.~\ref{fig:coarsess_control_comparison}, for example, the Arc de Triomphe can be approximated as a cuboid, leading the model to generate a generic office-like structure.
However, if the geometric prior includes an empty space in the middle, the model is able to infer a reasonable arch structure.
In the second row of Fig.~\ref{fig:coarsess_control_comparison}, the target building consists of a lower cuboid base and an upper tower.
From the simple cuboid, the model struggles to estimate the relative tower height.
When the prior explicitly specifies two stacked cuboids, SatSkylines successfully reconstructs the correct proportion between the upper and lower components. Similar improvements are also observed in additional real-world examples.

\begin{table}[]
\small

\makebox[\columnwidth][c]{
\centering
\begin{tabular}{l|ccc}
\hline
\multirow{2}{*}{Methods}   & \multicolumn{3}{c}{Geometry}                         \\
                           & IoU  $\uparrow$           & CD $\downarrow$ & F Score $\uparrow$         \\ \hline
Trellis   \cite{xiang2025structured}                  & 0.4415          & 0.0632           & 0.6295          \\
\textit{w finetune on Skylines-50K} & 0.6599          & 0.0273           & 0.8047          \\ \hline
\rule[-5pt]{0pt}{16pt}\textbf{SatSkylines}                & \textbf{0.9381} & \textbf{0.0168}  & \textbf{0.8684} \\
\textit{wo ss latent norm}          & 0.9300          & 0.0196           & 0.8598          \\ \hline
\end{tabular}
}
\caption{\footnotesize \textbf{Ablation Studies} include finetuning the image-only Trellis on our Skylines-50K dataset, and `\textit{wo ss latent norm}' refers to applying cosine geometric interpolation between the unnormalized $Z_{\mathcal{O}}$ and noise $\epsilon$.
}
\label{tab:ablation_studies}
\end{table}

Tab.~\ref{tab:coarseness_control} reports results on Skylines-50K using geometric priors of different coarseness. We define three levels of details (LODs):
LOD 0, a bounding box or simple cuboid;
LOD 1, one unique cross-section repeated along the building height;
LOD 2, two unique cross-sections.
IoU is highest for LOD 0 because its definition directly matches cuboid overlap with ground truth. 
Beyond IoU, however, higher LOD priors consistently lead to more accurate reconstructions, showing that SatSkylines benefits from increased prior detail while remaining robust to coarse inputs. 

\subsection{GPT Satellite Image Refinement}

As described in $\S$\ref{sec:realworld_pipeline}, in real-world settings, after obtaining the raw satellite image of a building, we refine it using `\textit{gpt-image-1}'.
This step is necessary since satellite imagery from Google Maps, Mapbox, or ArcGIS is often blurry and low-resolution.
Since the quality of the image prompt directly influences the fidelity of 3D building generation, improving clarity is essential.
As shown in Fig.~\ref{fig:gpt_refine_comparison} (left column), blurred satellite images lead to degraded 3D reconstructions. 
For example, gravel-like noise in the upper-left case and washed-out textures in the lower-left case.

Our pipeline employs `\textit{gpt-image-1}' with carefully designed text prompts to edit the original satellite image into a sharper, more detailed, and blur-free version. As shown in Fig.~\ref{fig:gpt_refine_comparison} (right column), 3D buildings generated from refined images exhibit clearer geometry and more realistic textures compared to those produced from the original blurred inputs.

\subsection{Ablation Studies}

We conduct ablation studies to validate the design choices of our method.
First, we finetune Trellis on the Skylines-50K dataset, which yields a noticeable improvement, reducing Chamfer Distance by 0.04. 
This demonstrates that our dataset facilitates the adaptation of a generalized 3D generation model to the satellite-view domain.

By adding our cosine geometric interpolation and SS latent normalization, we achieve an additional 0.011 reduction in Chamfer Distance, showing that the model successfully integrates coarse geometric priors into the generation process to produce more accurate building structures.
Finally, removing SS latent normalization results in a 0.003 drop in Chamfer Distance, indicating that aligning $Z_\mathcal{O}^{'}$ to have the same mean and standard deviation as pure Gaussian noise further improves generation quality.

\section{Conclusions}

In this work, we introduce \textbf{SatSkylines}, a 3D building generation approach that takes satellite imagery and coarse geometric priors.
To achieve this, we have developed \textbf{Skylines-50K}, a large-scale dataset of over 50,000 unique and stylized 3D building assets.
Built on top of our method, we further present a practical end-to-end pipeline that generates 3D building assets directly from GPS coordinates, enabling seamless application in real-world settings. Extensive evaluations indicates the effectiveness of our model and its strong generalization ability on both Skylines-50K and real-world scenarios.

{
    \small
    \bibliographystyle{ieeenat_fullname}
    \bibliography{main}
}

\clearpage
\newpage
\maketitlesupplementary
\setcounter{page}{1}
\setcounter{section}{0}

\begin{figure}[h]
\centering
\includegraphics[width=1\linewidth]{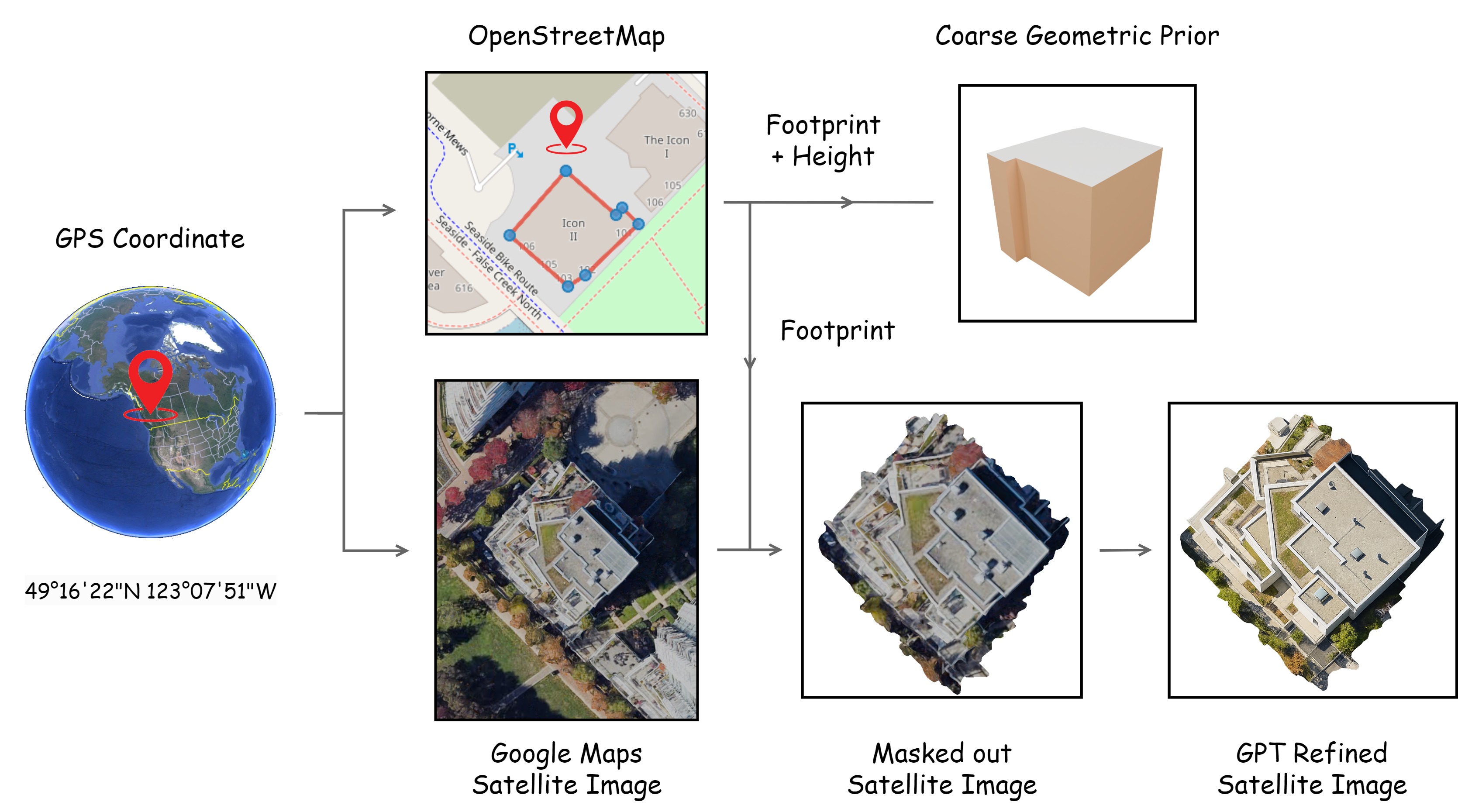}
\captionof{figure}{Real-world 3D building generation pipeline.}
\label{fig:real-world_pipeline}
\end{figure}

\section{Real-World End-to-End Pipeline}

In $\S$ \ref{sec:realworld_pipeline}, we have developed an end-to-end pipeline that requires only a geo-spatial bounding box of (min-lat, max-lat, min-lon, max-lon) to simplify the data preparation and building model generations. 
The overview of this pipeline is in Fig.~\ref{fig:real-world_pipeline}.
Specifically, from the bounding box information, it automatically retrieves 2D building footprints and height attributes from OpenStreetMap to generate coarse geometric priors, avoiding manual voxel creation. 
Satellite imagery from sources like Google Maps or Mapbox can also be combined with OpenStreetMap outlines to directly extract building masks without additional segmentation tools. 

\begin{figure}[h]
\centering
\includegraphics[width=1\linewidth]{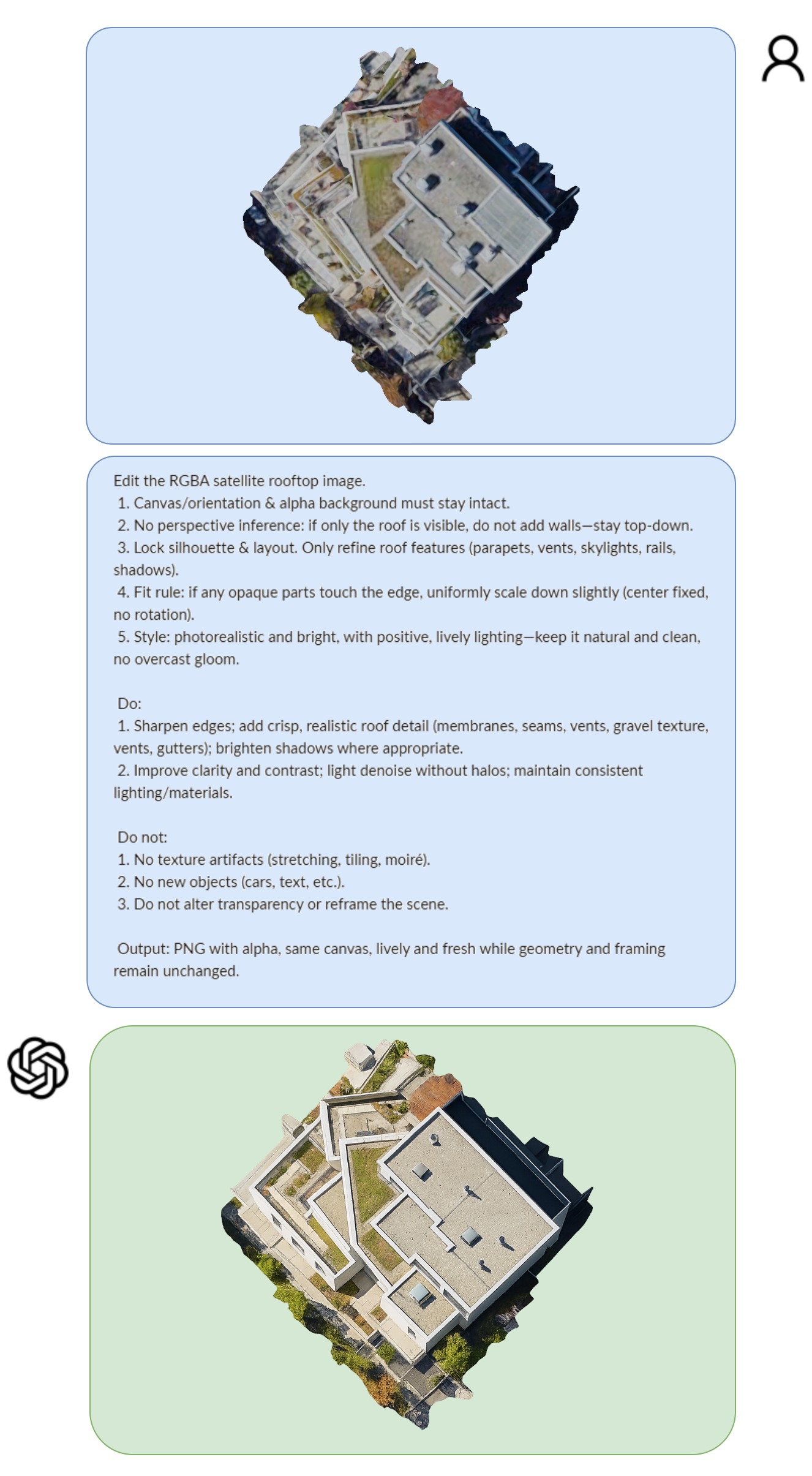}
\caption{An example of our GPT refinement process.}
\label{fig:gpt_refine_how_to}
\vspace{-1em}
\end{figure}

\textbf{GPT Satellite Image Refinement}.
Since satellite imagery is often blurry or low-resolution, our pipeline also performs a super-resolution step.
Although many super-resolution methods, such as MambaIRv2 \cite{guo2025mambairv2} and GSASR \cite{chen2025generalized}, have been proposed, they don’t consistently deliver satisfactory results for single-building satellite imagery.
We therefore adopt OpenAI’s `\textit{gpt-image-1}' for this task.
To be more specific, we send the raw satellite image together with a detailed, instruction-style prompt that specifies the input, the desired output, the operations to perform, and explicit “do-not” constraints.
An illustration of the entire refinement process can be found in Fig.~\ref{fig:gpt_refine_how_to}, which also includes the prompts designed for `\textit{gpt-image-1}'.

\section{More Visualizations}

\begin{figure*}[h]
\centering
\includegraphics[width=\linewidth]{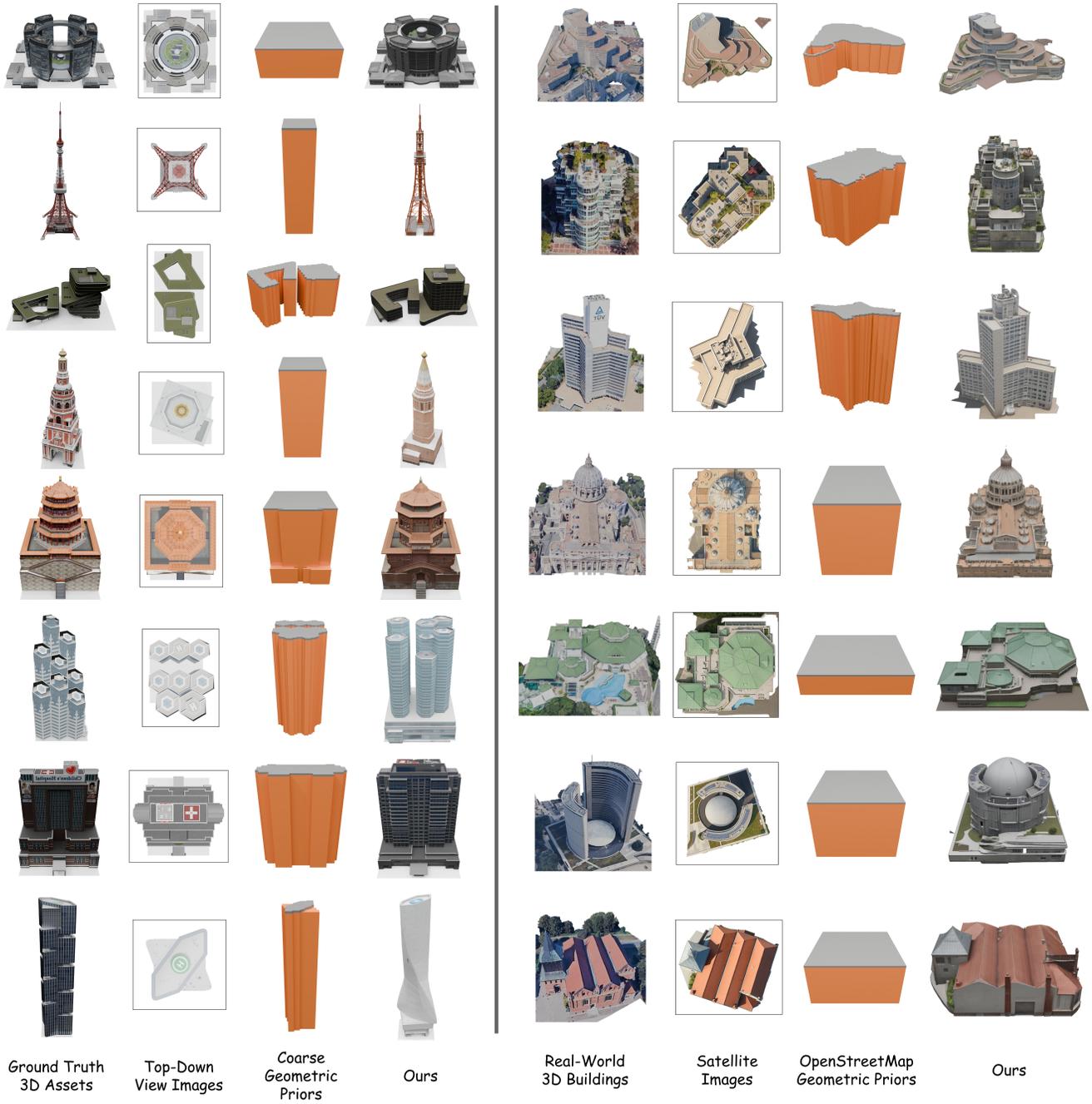}
\caption{More results generated by our SatSkylines. Left images are from Skyline-50K dataset, and right ones are sampled in real-world.}
\label{fig:more_results}
\end{figure*}

Figure~\ref{fig:more_results} shows additional 3D assets produced by our method: the left panel displays samples from the Skylines-50K test set, while the right panel presents randomly selected real-world cases. 
Given only coarse geometric priors, SatSkylines generates plausible buildings from satellite imagery. 
Because only a single overhead view is available, the synthesized assets are not expected to be identical to the ground truth; nevertheless, they still exhibit realistic geometry and a consistent, visually coherent appearance.

\section{Dataset}

As mentioned in $\S$\ref{sec:dataset_curation}, our Skylines-50K dataset contains $50,673$ 3D building assets, and we will release the dataset for future research.






\end{document}